\def\year{2022}\relax

\documentclass[letterpaper]{article} 
\usepackage{aaai22}  
\usepackage{times}  
\usepackage{helvet}  
\usepackage{courier}  
\usepackage[hyphens]{url}  
\usepackage{graphicx} 
\usepackage{amssymb}
\usepackage{natbib}  
\usepackage{caption} 
\DeclareCaptionStyle{ruled}{labelfont=normalfont,labelsep=colon,strut=off} 
\usepackage{algorithm}
\usepackage{algorithmic}

\usepackage{newfloat}
\usepackage{listings}
\usepackage{multirow}
\floatstyle{ruled}
\newfloat{listing}{tb}{lst}{}
\floatname{listing}{Listing}

\title{A Grammatical Compositional Model for Video Action Detection}
\author{
    Zhijun Zhang\textsuperscript{\rm 1},
    Xu Zou\textsuperscript{\rm 1}*,
    Jiahuan Zhou\textsuperscript{\rm 2},
    Sheng Zhong\textsuperscript{\rm 1},
    Ying Wu\textsuperscript{\rm 3}\\
}
\affiliations{
    \textsuperscript{\rm 1} Huazhong University of Science and Technology, Wuhan, Hubei, China
    \textsuperscript{\rm 2} Peking University, Beijing, China
    \textsuperscript{\rm 3} Northwestern University, Evanston, Illinois, United States\\
    zhangzhijun@hust.edu.cn, zoux@hust.edu.cn, jiahuanzhou@pku.edu.cn, zhongsheng@hust.edu.cn, yingwu@northwestern.edu\\
}

\usepackage{bibentry}

\begin{document}

\maketitle

\begin{abstract}
Analysis of human actions in videos demands understanding complex human dynamics, as well as the interaction between actors and context. However, these interaction relationships usually exhibit large intra-class variations from diverse human poses or object manipulations, and fine-grained inter-class differences between similar actions. Thus the performance of existing methods is severely limited. Motivated by the observation that interactive actions can be decomposed into actor dynamics and participating objects or humans, we propose to investigate the composite property of them. In this paper, we present a novel Grammatical Compositional Model (GCM) for action detection based on typical And-Or graphs. Our model exploits the intrinsic structures and latent relationships of actions in a hierarchical manner to harness both the compositionality of grammar models and the capability of expressing rich features of DNNs. The proposed model can be readily embodied into a neural network module for efficient optimization in an end-to-end manner. Extensive experiments are conducted on the AVA dataset and the Something-Else task to demonstrate the superiority of our model, meanwhile the interpretability is enhanced through an inference parsing procedure.
\end{abstract}

\section{Introduction}

\begin{figure}
    \centering
    \includegraphics[width=1\linewidth]{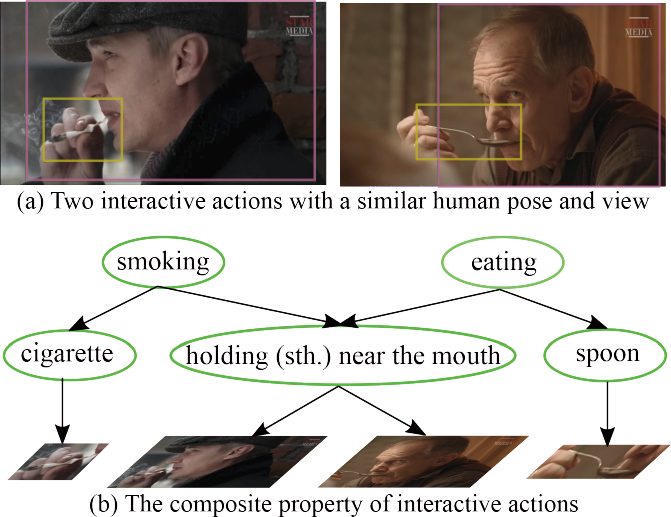}
    \caption{How do you differentiate two actions with a similar human pose and view in (a), ``smoking" or ``eating"? 
    Humans can easily infer the consequences with different interacting objects (\textit{cigarette} or \textit{spoon}). We observe that interactive actions have composite property, that they can be decomposed into distinct components, human dynamics, and interacting entities, like (b). We propose a grammatical compositional model to exploit the property for action detection.}
    \label{fig:challenges}
\end{figure}

How to capture interactions between humans and context is of central importance in the action detection task~\cite{zhang2019structured,pramono2019hierarchical}, since most actions are performed with interaction to objects or other people in daily life, \textit{e.g.}, \textit{holding (sth.)} or \textit{talking to (sb.)}. 
However, action detection still remains challenging due to its large intra-class variations from diverse human appearance or object manipulations, and small inter-class differences between similar actions as shown in Figure \ref{fig:challenges}(a).

To alleviate the above mentioned issues, two categories of algorithms have been proposed to model interactions \cite{zhang2019structured,tomei2019stage,yao2010grouplet} in action detection. One class \cite{sun2018actor,pramono2019hierarchical,girdhar2019video,wu2020context} seeks to automatically localize relevant contextual information to support action representations using relation network modules \cite{santoro2017simple} or self-attention mechanisms~\cite{vaswani2017attention}.
Such approaches, however, do not model objects explicitly which simply ignore the structures of interactive actions. Another kind of methods \cite{zhang2019structured,tang2017towards} exploit the structures of interactions explicitly and directly. But they either build a single-layered relation graph \cite{zhang2019structured} which can not model complex interactions with objects and humans simultaneously, or design sophisticated modules making it difficult to understand intuitively \cite{tang2020asynchronous}.

To tackle these issues, we account for the intrinsic compositionality of interactive actions. 
As shown in Figure \ref{fig:challenges} (b), two actions, \textit{smoking} and \textit{eating}, can be decoupled into similar actor dynamics (\textit{holding (sth.) near the mouth}), and different interacting objects (\textit{cigarette} and \textit{spoon}). 
These decomposed components can be utilized to generate various configurations of actions, where the sharing of reusable elements (actor dynamics) could reduce the complexity of the action representation with huge intra-class variations.

Grammatical model \cite{si2011unsupervised,zhu2007stochastic} is a natural choice for modeling the compositionality due to the inherent syntactic grammar of actions (\textit{subject-verb-object}).

Previous grammatical models \cite{li2019aognets,tang2017towards} often address general And-Or grammar operations \cite{geman2004probability,zhu2007stochastic} to empower the DNNs with the composite properties, or focus on the explicit hierarchical grammar modeling of human poses \cite{tang2018deeply}.

In contrast, we investigate the compositional hierarchy of interactive actions in videos. However, learning a structural grammatical model for action detection is not easy. Firstly, one human body could perform multiple actions simultaneously. \textit{e.g.}, ``sit while eating'' or ``dance while singing''. Simple model exclusive primitive actions would produce inaccurate results. Secondly, due to unknown locations of participating objects or humans, leaf nodes of the model is hard to choose visual attributes from videos.

\begin{figure*}
    \centering
    \includegraphics[width=1\linewidth]{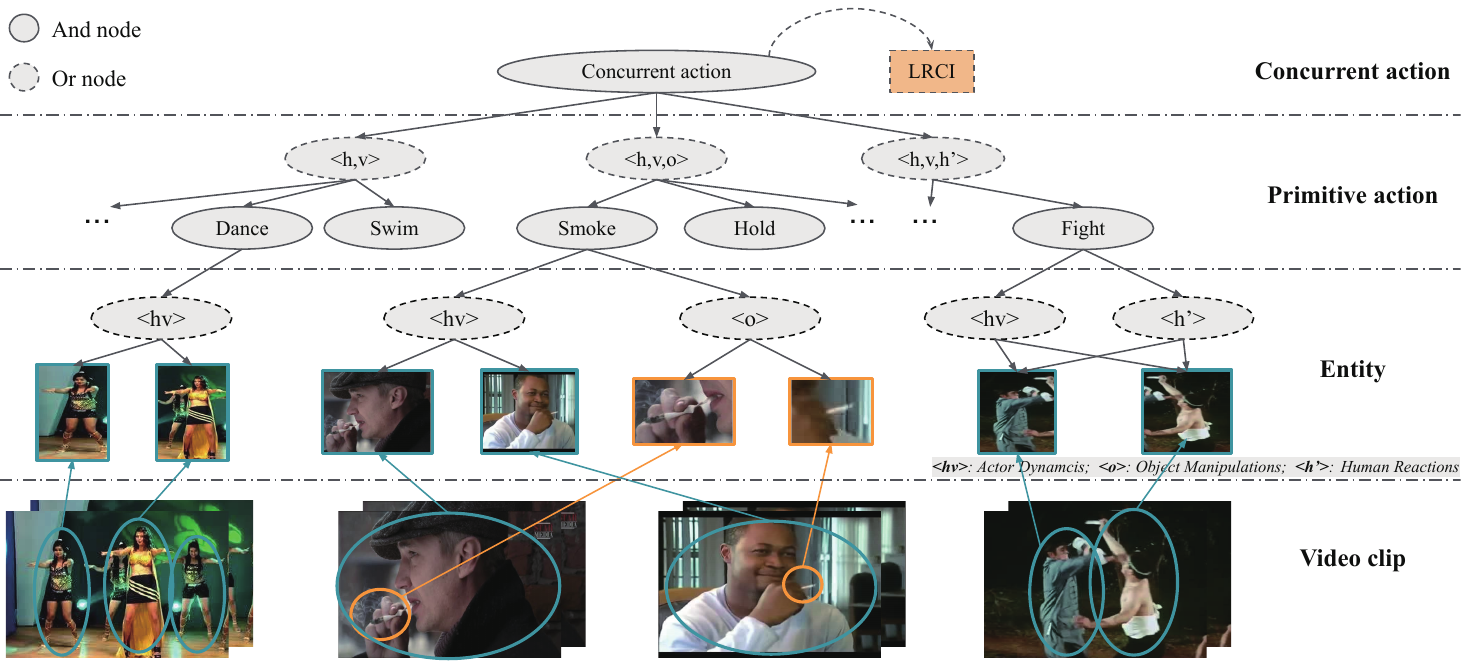}
    \caption{Overview of our \textbf{Grammatical Compositional Model (GCM)} for human action detection. At the top level, \textbf{Concurrent action} nodes, defined as multiple actions performed by one person, contain three interactive types of actions: \textit{human body movements} $<h,v>$, \textit{human-object interactions} $<h,v,o>$, and \textit{human-human interactions} $<h,v,h'>$. 
    \textbf{Primitive action} nodes (\textit{i.e.}, single actions), alternatively selected by interactive types, are in turn the composition of entities. \textbf{Entity} nodes, including \textit{actor dynamics} $<hv>$, \textit{object manipulations} $<o>$, and \textit{human reactions} $<h'>$, are various modalities of spatiotemporal features extracted from video clips based on human and object detection results. \textbf{LRCI} is the Long-Range Contextual Information, detailed in Figure \ref{fig:long-range} and Sec. \ref{lrci}.}
    \label{fig:overview}
\end{figure*}

To the goal of compositional structure modeling of actions in videos, 
we propose a novel Grammatical Compositional Model (GCM) for action detection with the following characteristics: 1) Concurrent actions are defined as non-exclusive compositions of primitive actions; 2) Primitive actions are defined by compositions of entities based on interactive types; 3) Entities are defined as spatial-temporal configurations of actors dynamics, object manipulations and human reactions grounded from the video data. The video is encoded into multiple person and object proposals as leaf nodes, and infer the action classes of each person through bottom-up detections in our compositional model, as shown in Figure \ref{fig:overview}. Owing to the model-agnostic merit of our proposed approach, we could readily embody our method into existing neural network modules to fully take advantage of the powerful expression ability of DNNs.

Moreover, we also address the long-range temporal modeling problem \cite{wu2019long,tang2020asynchronous} in videos to complement action predictions from short video clips in our unified model. 

In summary, the contributions of this paper are as follows:

\begin{itemize}
    \item We propose a novel Grammatical Compositional Model (GCM) for video action detection, which investigates the compositional hierarchy of interactive actions based on the And-Or grammar. To the best of our knowledge, we are the initial work to learn grammar models for action detection via deep networks.
    \item Our model generates enormous configurations of actions hierarchically and builds spatial and long-range temporal relationships jointly in a simple and efficient~manner.
    \item Our proposed GCM is plug-and-play and can be readily integrated into a deep network module, which harnesses the compositionality of grammar models and the powerful expressive ability of DNNs for better performance.
    \item We conduct various extensive experiments on the challenging AVA dataset and another compositional action recognition task \cite{materzynska2020something} to validate the effectiveness of our model, and show that the interpretability is enhanced through grammatical parsing.
    
\end{itemize}

\section{Related Work}

\noindent\textbf{Spatio-temporal action detection} is an active research area. Many approaches \cite{singh2017online,peng2016multi,kopuklu2019you} focus on generalizing remarkable deep learning based object detection algorithms to spatio-temporal action detection, such as Faster-RCNN \cite{ren2015faster}, SSD \cite{liu2016ssd}, YOLO \cite{redmon2016you}. Others \cite{song2019tacnet,li2018recurrent} target sequential modeling by utilizing the recurrent model \cite{schuster1997bidirectional} or LSTM \cite{hochreiter1997long} to capture the dynamics of human actions. With the presentation of dataset AVA, which involves many complex interactive actions, more and more works \cite{sun2018actor,zhang2019structured,tang2020asynchronous} aim at modeling the relationships between actors and context for boosting action detection. Our work follows the research~line.

\noindent\textbf{Interaction modeling} has been studied for a certain time.
Many approaches \cite{gkioxari2018detecting,chao2018learning,ulutan2020vsgnet,xu2019learning,xiao2019reasoning} model the relationships between human and objects in still images. However, they fail to disambiguate some actions, such as \textit{pick-up} or \textit{pick-down} due to the lack of motion. And some works seek to utilize the interaction information for fostering video recognition task \cite{wang2018videos,herzig2019spatio,chen2019graph,li2020adaptive,ma2018attend} and spatio-temporal action detection \cite{sun2018actor,tang2020asynchronous}. Instead of modeling the interactive actions as a relation graph \cite{wang2018videos,zhang2019structured} or an interaction block \cite{tang2020asynchronous}, we are the initial work to implement grammatical compositional model in interaction modeling.

\noindent\textbf{Compositional model} has been explored in many vision tasks \cite{tang2017towards,li2019aognets}, involving object detection  \cite{song2013discriminatively,si2013learning}, human pose estimation \cite{tang2018deeply,duan2012multi} and activity recognition \cite{si2011unsupervised,wang2014cross,liang2013learning,li2019hake}. 
In contrast to their applications, we focus on the human-centric action detection problem by analyzing the composite property of actions, and do not require additional annotations of objects \cite{materzynska2020something} and spatiotemporal relation graphs \cite{ji2020action}, alleviating labeling burdens.

Several works focus on the fusion of compositional/grammatical models with DNNs. \cite{tang2017towards} and \cite{li2019aognets} study a general grammatical compositional design of deep network architectures, and validate their effectiveness on image classification \cite{russakovsky2015imagenet} and object detection \cite{lin2014microsoft} tasks. 

Following them, we embody our compositional model into a network module to maximize performance.

\noindent\textbf{Long-term video understanding} with modern CNN has been studied by a few works \cite{wu2019long,tang2020asynchronous,wang2016temporal,zhou2018temporal}. The long-range features could provide supportive information or contextual guidance for improving video models. Different from them, our approach seeks to fuse the long-term video information into our hierarchical compositional structure, in which the long-term semantics are the compositions of concurrent actions during a relatively long period. 
\section{Grammatical Compositional Model}

We first make a brief introduction of our grammatical compositional model (Sec. \ref{overview}). The overview of our GCM is illustrated in Figure \ref{fig:overview}, where long-range contextual information, aiming at complementing short-term action inference, is detailed in Figure \ref{fig:long-range}. We then describe the node representation and scoring function of each layer respectively (Sec. \ref{concurrent} - \ref{lrci}), consisting of concurrent actions, primitive actions, entities, and long-range contextual~information.

\subsection{Overview}\label{overview}

Our grammatical compositional model is defined on a hierarchical AND-OR graph, following the image grammar framework proposed by Zhu \cite{zhu2007stochastic}. It is characterized by a 3-tuple $\mathcal{G}=(\mathcal{V},E,\Theta)$, where $\mathcal{V} = \mathcal{V}_{And} \cup \mathcal{V}_{Or} \cup \mathcal{V}_{L}$ represents the set of nodes, consisting of three subsets of And-nodes, Or-nodes and Leaf-nodes respectively. And-nodes encode the composition of parts, and Or-nodes account for alternative configurations. $E$ is the set of edges organizing all nodes, and $\Theta$ is the set of parameters.

\begin{figure}
    \centering
    \includegraphics[width=1\linewidth]{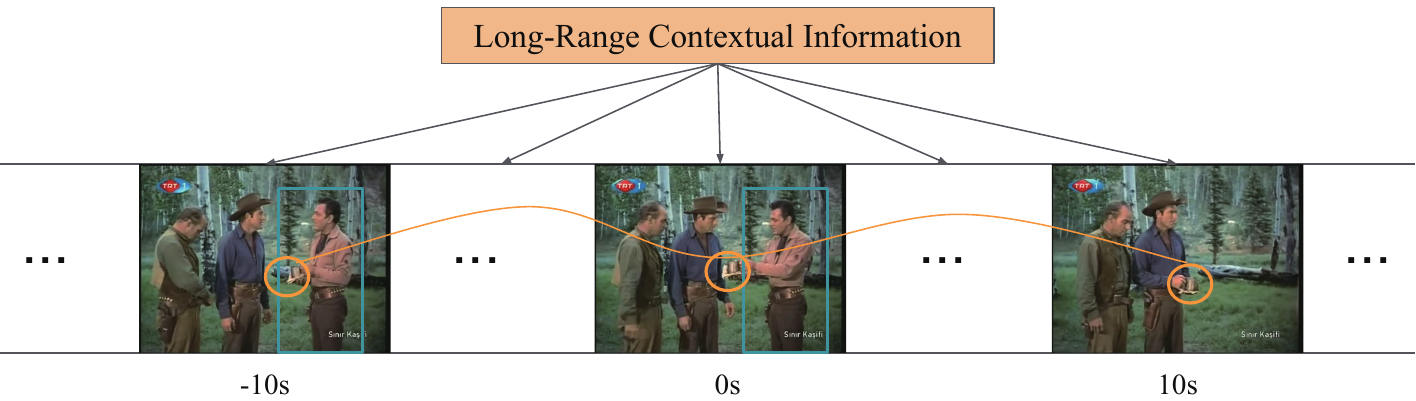}
    \caption{Overview of the Long-Range Contextual Information (LRCI), which is a composition of several contiguous video clips from a long video. This information could provide supportive cues (such as the cup location change in the yellow circle) to guide the model in better understanding the present actions. (AVA ground-truth: ``give (sth.) to (sb.)")}
    \label{fig:long-range}
\end{figure}

The proposed GCM, illustrated in Figure \ref{fig:overview}, organizes domain knowledge in a hierarchical manner. Previous works \cite{sun2018actor,gu2018ava} often process each action co-performed by one person individually, which may ignore the semantic relationships among actions that happened together. To address the issue, we introduce \textit{Concurrent actions}, $a \in A$, which are defined as multiple actions simultaneously performed by one human body \cite{wei2013concurrent}. 
These concurrent actions are certainly the composition of \textit{Primitive actions}, $r \in R$. Since there exist several different interactive manners among the primitive actions, we divide them into three interactive types of actions following \cite{gu2018ava,zhang2019structured,tang2017towards}, \textit{i.e.}, human body movements ($<h,v>$), interactions with objects ($<h,v,o>$), and interactions with humans ($<h,v,h'>$). As illustrated in Figure \ref{fig:challenges}, 
the primitive actions can be decomposed into distinct \textit{Entities}, $e \in E$. Here we involve three types of entities: actor dynamics $<hv>$, object manipulations $<o>$ and human reactions $<h'>$. Each entity represents various appearance and motion modalities of humans and objects from video clips.

To make action predictions more accurate, we additionally introduce the \textit{long-range contextual information}, $l \in L$, detailed in Figure \ref{fig:long-range}, which is consisted by actions over a long-term temporal duration. This information
could provide 
supportive cues to complement the action inference from short video clips, and disambiguate some noisy predictions. 
We extend our GCM model by integrating this node $L$ with the concurrent action composition $A$ to construct a more exhaustive action representation.

\begin{figure}
    \centering
    \includegraphics[width=1\linewidth]{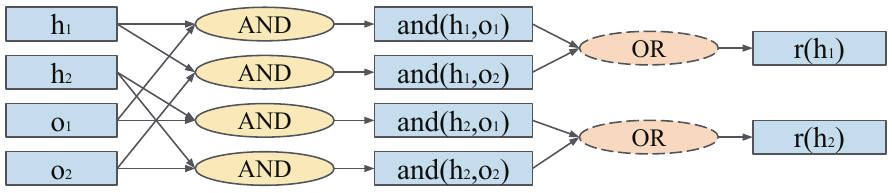}
    \caption{An example of primitive action inference ``$r$'' with the entity features of two human and object proposals via And-Or operations. We firstly compose pair-wise human and object interactions by And nodes, and then select the optimal pairs for each person by Or nodes.}
    \label{fig:Primitive Action Inference}
\end{figure}

\subsection{Concurrent Action Nodes}\label{concurrent}

Concurrent actions are recognized as the multiple actions simultaneously performed by one human body, which has been defined in \cite{wei2013concurrent}. 

Modeling those actions together could leverage the inherent correlation of them, \textit{e.g.}, \textit{working on the computer} often happens with \textit{sitting}, not \textit{standing}.
We consider the concurrent action $A \in \mathcal{V}_{And}$ as a composition of three groups of primitive actions based on their interactive types, \textit{i.e.}, \textit{human body movements}, \textit{human-object interactions}, and \textit{human-human interactions}. Following \cite{tang2017towards}, we use a scoring function, defined as the negative energy of probability function with the Gibbs distribution, to specify the model. The scoring function at the concurrent action node $A_0$ is defined as:

\begin{equation}
    S_A(a_0,\theta) = \sum_{r \in ch(A)} \mathop{max}\limits_{\omega_{r}} [S_R(\omega_r,\theta)], 
\end{equation}

\noindent where $r$ denotes the primitive actions. The state variable $\omega_r$ encodes the interactive type of each primitive action, involving $<h,v>$, $<h,v,o>$, and $<h,v,h'>$, which is often pre-decided by action categories. The term $S_R$ describes the scoring function of primitive actions $r$. The ``$\sum$'' and ``max'' represent And and Or node operations respectively, demonstrating the compositions of primitive actions, and alternative selections for their interactive types respectively.

\subsection{Primitive Action Nodes}\label{primitive}
For a primitive action $R \in \mathcal{V}_{AND}$, it contains ``who" performs the action, \textit{i.e.}, actors $h$, and ``what" component is interacting with, either objects $o$ or humans
$h'$. These components describe people involved, and related objects in a single action. Thus we decompose a primitive action into distinct entities based on its interactive type as shown in Figure \ref{fig:overview}.  
For example, a human-object interaction $<h,v,o>$, \textit{e.g.}, ``\textit{smoking}", is decomposed into actor dynamics, termed as ``$hv$", and participating objects manipulation, ``$o$" (\textit{e.g.},  \textit{cigarettes or pipe}). And a human-human interaction $<h,v,h'>$, \textit{e.g.}, ``\textit{fight}" is decoupled into actor dynamics ``$hv$" and interacting human reactions ``$h'$". 

However, the problem of applying simple decomposition is that we do not know ``what'' component is interacting with. To handle this issue, we account for embedding the interacting component options in the state variable of entity nodes, \textit{i.e.}, $\omega_e \in \{o_1,o_2,...,o_{N_o}, h_1, h_2, ..., h_{N_h} \}$, and optimize it by maximizing the scoring function at primitive action nodes $R$, which is defined as:

\begin{equation}
    S_R(\omega_r, \theta) = \mathop{max}\limits_{\omega_{e}} \sum_{e \in ch(r)}  [S_E(\omega_e,  \theta)],
\end{equation}

\noindent where $S_{E}$ is the scoring function of entity nodes. 

$\omega_e$ is the state variable of entity nodes, determining the selection of region proposals. 
We use And nodes to compose primitive action proposals and then alternatively select primitive actions for each person by Or nodes. An example of primitive action inference is illustrated in Figure \ref{fig:Primitive Action Inference}.

\subsection{Entity Nodes}\label{entity}

The entity nodes contain human dynamics $hv$, object manipulations $o$ and human reactions $h'$. Each entity node captures the visual appearance and motion features extracted from video clips based on human and object detection results. 
To fill the gap between grammatical symbols and powerful CNN feature representation, we use 3DCNN networks \cite{feichtenhofer2019slowfast} to generate spatiotemporal features and implement object detection networks \cite{ren2015faster,massa2018mrcnn} to capture the localization of humans and objects due to their strong expressive ability.

Aiming at providing sufficient information for action detection in both appearance and motion space,
we define the scoring functions of three types (human dynamics, object manipulations and human reactions) of entity nodes as:
\begin{eqnarray}
    S_{E}(\omega_e, \theta | e \in hv) &=& f(p_h,  \theta)\\ 
    S_{E}(\omega_e, \theta | e \in o, \omega_e = (o_j)) &=& f(p_{o_j},  \theta)\\
    S_{E}(\omega_e, \theta | e \in h', \omega_e = (h'_j)) &=& f(p_{h'_j}, \theta),
\end{eqnarray}

\noindent where $p_h, p_{o_j}, p_{h'_j}$ are the localization of actor, interacting object and human proposals in the keyframe inferred by Faster-RCNN network \cite{ren2015faster}, and $f(p,\theta)$ is the feature extraction function by 3DCNN based on region proposals $p$. The powerful networks could provide rich information for supporting action prediction.
It is worth noting that we rely on off-the-shelf human and object detectors to generate proposals, so that they can filter many irrelevant background information and allow our model to put more emphasis on the discriminative parts in the composition.

\subsection{Long-range Contextual Information}\label{lrci}

Action detection inferred from short video clips often causes some ambiguous results due to motion blur or scene switching in movies. 
To make action detection more accurate, we further model the relationships between what is happening in the present and actions that occurred distant in time. As illustrated in Figure \ref{fig:long-range}, if we observe the location changes of the cup in the frames from \textit{-10s} to \textit{10s}, it would be much easier to predict the action ``give (sth.) to (sb.)" performed by the person in the present.

To address the issue, we introduce long-range contextual information $L$ as the set of actions during a relatively long period of time. 
We consider the information as the supportive cues for our action inference, and compose it with our concurrent action compositions to make a more accurate representation. We rewrite the scoring function of concurrent action nodes $A^*$ for aggregating them as follows:

\begin{equation}
    S_{A^*}(a_0, \theta) = S_A(a_0, \theta) + S_L(\omega_l,\theta),
\end{equation}

\noindent where $S_A$ is the scoring function from the composition in (1), and $S_L$ is the scoring function of long-range contextual information. We consider the function $S_L$ should automatically select the most supportive information. Thus we formulate $S_L$ similar with primitive action inference $S_R$ in (2), where the scoring function is as:

\begin{equation}
    S_L(\omega_l,\theta) = \mathop{max}\limits_{\omega_t} \sum_{t \in ch(l)}  [S_A(a_t, \theta)],
\end{equation}

\noindent where $a_t$ is the concurrent action in the $t$-th video clip. $\omega_t$ is the state variable, denoting the concurrent action options in long-range contextual information, \textit{i.e.} $\omega_t \in \{ a_{-T}, ..., a_{-1}, a_{1}, ..., a_{T}\}$, where we set $T = 30$. We use And and Or node operations to select the supportive information from these concurrent actions for boosting the current action detection.
Since simultaneously calculating long-range information is nearly unachievable due to limited GPU memory and computational sources, we store and asynchronously update concurrent action scores $S_A$ similar with \cite{tang2020asynchronous}.

\section{Inference via DNN Modules}

\begin{figure}
    \centering
    \includegraphics[width=1\linewidth]{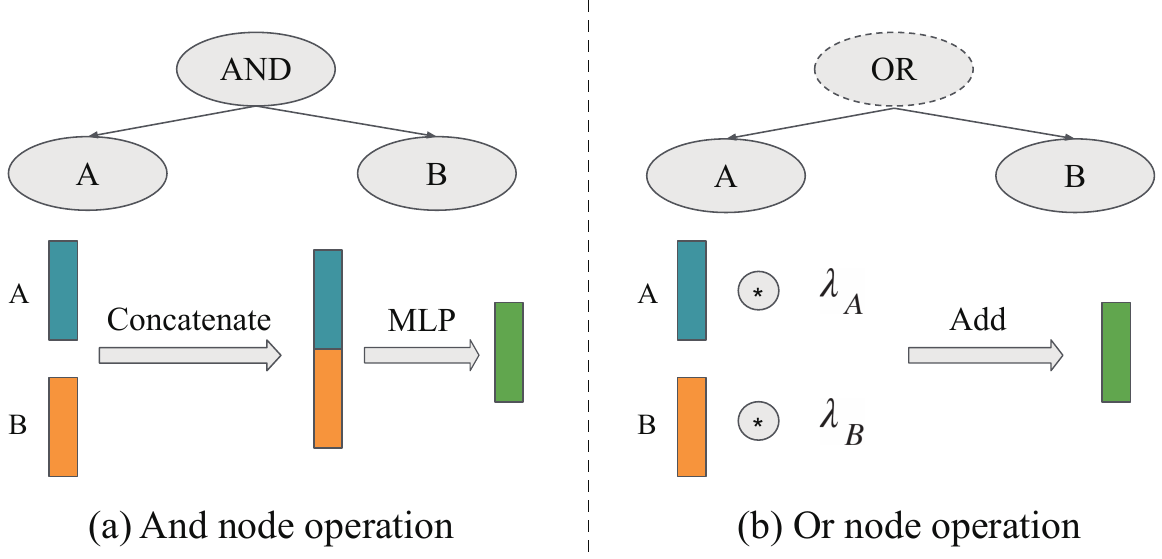}
    \caption{An illustration of two types of node operations in the GCM network module with two components.}
    \label{fig:network}
\end{figure}

Empirical studies \cite{tang2017towards,li2019aognets} demonstrate that And-Or graphs can be embodied into convolutional neural networks to achieve better performance. This inspires us to embody our GCM in a network module, to bridge the gap between the composite property of grammar models and the powerful expression ability of DNNs.

In our model, there are two types of node operations: ``And" operation and ``Or" operation, which correspond to ``$\sum$" and ``$\mathop{max}$" in the former scoring functions. As illustrated in Figure \ref{fig:network}, we realize them by some simple operations in neural networks:

\begin{itemize}
    \item ``And" operation is a composition rule $S \rightarrow [A \cdot B]$, which decomposes the parent node $S$ into distinct child nodes $A$ and $B$. It is performed by a concatenate operation plus a two-layer MLP in the module, \textit{i.e.}, $f_{out} = g([A,B])$;
    \item ``Or" operation represents alternative ways of a symbol $S \rightarrow A|B|C$, and we utilize weighted add to select elements, \textit{i.e.}, $f_{out} = \lambda_1 A + \lambda_2 B + \lambda_3 C, s.t. \sum_i \lambda_i = 1$, where $\lambda$ is the weight of each component, calculated by a MLP with the softmax function on input elements.
\end{itemize}

Our GCM model is inferred through a bottom-up scheme, the leaf/entity nodes of the graph are obtained by spatiotemporal features based on human and object detection results. And the following nodes are calculated based on their scoring functions detailed in the former. The final action class prediction is obtained by a 2-layer MLP on the scoring maps of the root node.
The constructed module can be connected with any feature generators to form an end-to-end network for efficient~optimization. 
\section{Experiments on AVA}

We evaluate our proposed model on the AVA dataset \cite{gu2018ava}, which contains 235 training videos and 64 validataion videos generated from 80 different actions. 

Compared with traditional action detection datasets, \textit{i.e.}, UCF101-24 \cite{soomro2012ucf101} and JHMDB \cite{kuehne2011hmdb}, we address the interaction issues in the AVA dataset for two reasons: a) AVA exhibits high complexity in human-context interactions, and b) the distribution of action classes in AVA is more close to our daily life, rather than artificially balanced datasets.

\subsection{Implementation Details}

The proposed compositional model is performed as a network module, which can link to any CNN video representation with entity proposals. We describe the representation in two aspects: a) the localization of human and object proposals in keyframes; b) the spatiotemporal features extracted from video clips.

\noindent\textbf{Entity proposals localization.} 
In the experiments, we localize human proposals in keyframes using Faster R-CNN \cite{ren2015faster} with a ResNeXt-101-FPN \cite{lin2017feature,xie2017aggregated} backbone, and localize object proposals by Mask R-CNN \cite{massa2018mrcnn}, as same as \cite{tang2020asynchronous}. The human detection achieves 93.9 AP@50 on the AVA validation set. We select human proposals with IoU $\textgreater$ 0.8 and top-5 object proposals as our leaf nodes.

\noindent\textbf{Video spatiotemporal features.} We apply the state-of-the-art Slowfast \cite{feichtenhofer2019slowfast} network with the ResNet-50 structure as our backbone to generate video spatiotemporal features. The network is pre-trained on Kinetics-700 \cite{carreira2017quo} for the video classification, and then fine-tuned on AVA datasets with an ROI head. 
The inputs of the backbone are 32 RGB frames sampled with a temporal stride of 2 spanning 64 frames. Clips are scaled to 256 on the short side. 
It provides powerful entity representations in videos, where the leaf symbols with dimension $d=2304$ are extracted via ROI align and temporal average pooling operations.

\noindent\textbf{Implementation of network module.} Given the features of leaf nodes, $f_h \in \mathbb{R}^{n_h \times d}$, $f_o \in \mathbb{R}^{n_o \times d}$, we first perform compositions of each layers, and then integrate long-range contextual information to construct the final scoring map. The dimension of scoring maps of primitive actions and concurrent actions is 1024. And the entity nodes are implemented by a reducing dimension of leaf symbols, where $d_e = 512$. The parameter $T$ in long-range contextual information is set as 30 to capture one-minute video information. Finally, we predict the action label from the last layer by a 2-layer~MLP.

\noindent\textbf{Training and inference.} Considering the training time and memory limit, we divide the training process into three stages. Firstly, we only train the backbone network on the AVA dataset, \textit{i.e.}, Slowfast network \cite{feichtenhofer2019slowfast}. The network is trained using the SGD algorithm with batch size 16 on 4 GPUs and learning rate 0.05, reduced by a factor 10 at each 5 epochs. 

We store all features of human and object proposals inferred from the backbone in the disk. Secondly, using all pre-computed features of proposals, we train a pure GCM module by Adam algorithm, with a learning rate of $10^{-4}$ at the first 3 epochs and $6.5 \times 10^{-5}$ in the latter. We apply Dropout \cite{hinton2012improving} on the last compositional layer with a ratio of 0.5. Finally, we aggregate the backbone and GCM module as an end-to-end network, and fine-tune them together to achieve better performance. In all three stages, we use ground-truth human bounding boxes for training, and detected human boxes for inference with a confidence score larger than 0.8. To avoid the influence of human detection error, we extend the bounding boxes slightly with a ratio of 0.2 in both training and testing.

\begin{figure}
    \centering
    \includegraphics[width=\linewidth]{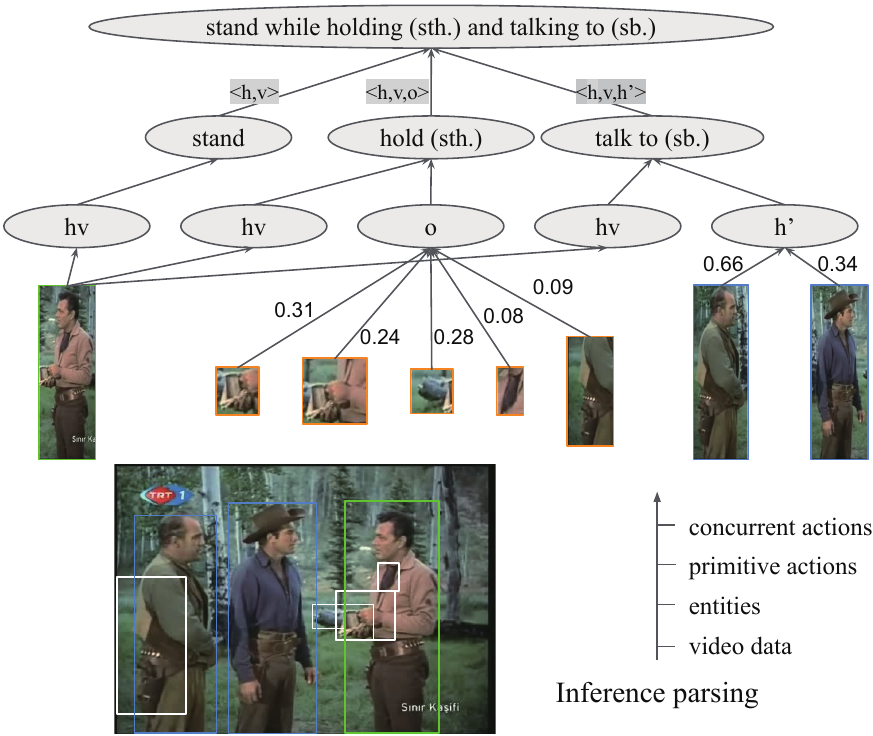}
    \caption{An example of the inference parsing result. Different entity proposals support the action inference with learned weights in Or nodes. Notice the ``cup'' object contributes the most to the primitive action ``hold (sth.)''.}
    \label{fig:parsing}
\end{figure}

\subsection{Experimental Results}

\begin{table}
    \centering
    \resizebox{0.85\linewidth}{14mm}{
    \begin{tabular}{c|c|c}
    \hline
    \hline
        ~ & Top-5 categories & Gain (mAP) \\
        \hline
        1 & listen to (\textit{e.g.}, a person) & +12.62 \\
        2 & cut & +11.14 \\
        3 & listen (\textit{e.g.}, to music) & +9.39 \\
        4 & play musical instrument & +9.28 \\
        5 & hit (a person) & +8.91 \\
        \hline
        \hline
        
    \end{tabular}}
    \caption{The top-5 categories improved by our approach on AVA, compared to the baseline. Our approach gains a large improvement on interactive actions.}\label{tab:gain}
    \label{tab:my_label}
\end{table}

\noindent\textbf{Inference parsing.} To show the interpretability of our model, we first give an example of inference parsing from a single clip in Figure \ref{fig:parsing}. It is intuitive that the concurrent action inference originated from the video data could give a parsing result for each object and human proposal by a bottom-up scheme. The weights of all children nodes decomposed by Or nodes are illustrated, showing the supportive scores for the action predictions.

\begin{figure}
    \centering
    \includegraphics[width=1\linewidth]{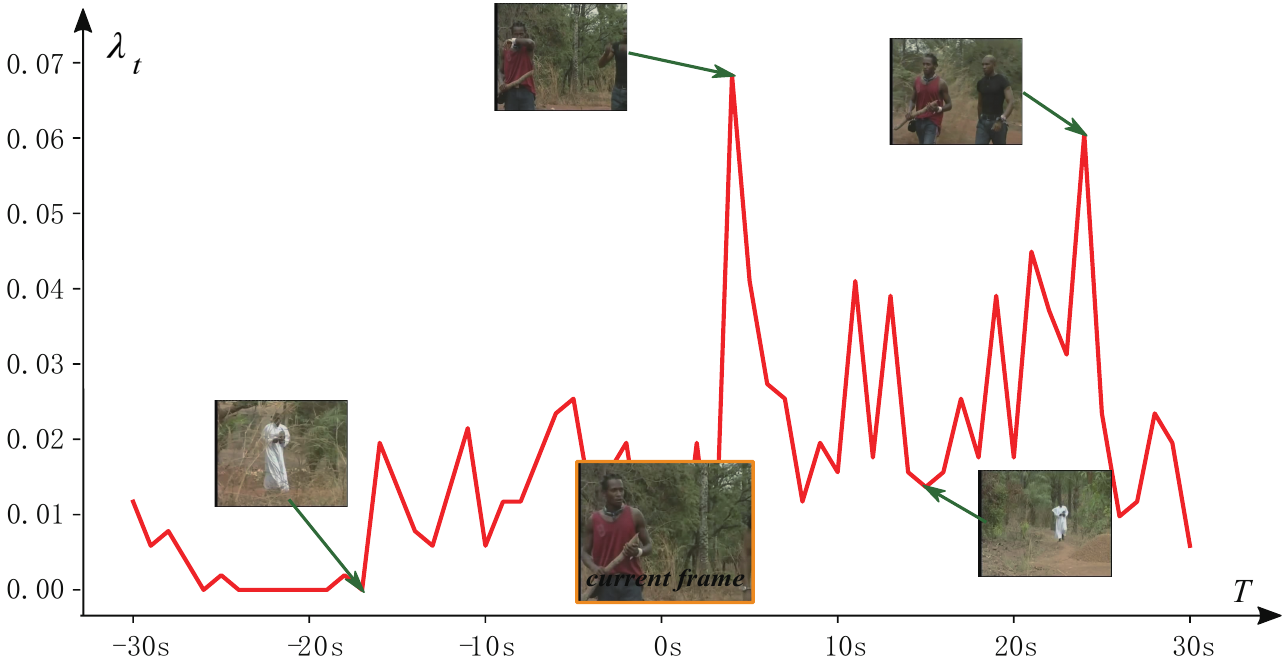}
    \caption{Inference parsing on the long-range contextual information (1 min). Notice that frames with higher scores are more related to the current frame rather than lower ones.}
    \label{fig:supportive}
\end{figure}

Furthermore, we also visualize the supportive scores $\lambda_t$ of each second in Figure \ref{fig:supportive} to validate our long-range contextual information modeling. The value of scores indicates the degree of contribution for the current frame. It is seen that even though long-term videos have much noisy and irrelevant information due to scene switching, our model could automatically capture the most related cues.

\begin{table}
\centering
\resizebox{\linewidth}{15mm}{

\begin{tabular}{c|c|c}
\hline
\hline
& \textbf{Model} & \textbf{mAP}  \\
\hline
1 & Baseline & 26.74  \\ 
2 & Primitive Action Layer & 28.59  \\ 
3 & Concurrent Action Layer & 28.78\\
4 & Concurrent Action Layer with LRCI & 29.76 \\
5 & Concurrent Action Layer with LRCI * & \textbf{30.02} \\
\hline
\hline
\end{tabular}}
\caption{Evaluation of each layer predictions in our model. LRCI is the Long Range Contextual Information, and * means the module is fine-tuning with the backbone jointly.}
\label{tab:layer}
\end{table}

\noindent\textbf{Does interaction modeling really help?} To show the effectiveness of our modeling for interactive actions, we report the top-5 categories improved by our approach on AVA compared to the baseline in Table \ref{tab:gain}. As demonstrated, the major improvements focus on interactive action categories (\textit{e.g.}, \textit{listen to (sb.) or cut (sth.)}), which are exactly what we research on. It is worth noting that the third category \textit{listen (\textit{e.g.}, to music)} and the fourth category \textit{play musical instrument} have significant semantic correlations, and both of them exhibit a large improvement.

\noindent\textbf{How much each layer helps?} To validate our hierarchical layer design, we perform an ablation study based on pre-computed features of ResNet-50 Slowfast in Table \ref{tab:layer}.
As demonstrated, each layer has shown a stable improvement of performance against the previous layer, which verifies the effectiveness of our hierarchical design. The most significant improvement comes from the Primitive Action Layer, obtaining a boost of 6.9 \%. This enhancement demonstrates the importance of entity compositions for interactive actions. Combining long-range information also gains an appreciable improvement from 28.78 mAP to 29.76 mAP, indicating the support of long-range~modeling.

\begin{table}
\centering
\footnotesize
\resizebox{\linewidth}{22mm}{

\begin{tabular}{c|c|c}
\hline
\hline
\textbf{Approach}   & \textbf{Backbone} & \textbf{mAP}  \\
\hline
AVA\cite{gu2018ava}  & I3D & 15.6  \\
ACRN\cite{sun2018actor}  & S3D & 17.4  \\
STEP\cite{yang2019step}  & I3D & 18.6  \\
Relation graph\cite{zhang2019structured}  & R50-NL & 22.2 \\
VAT\cite{girdhar2019video}  & I3D & 24.9 \\
Slowfast\cite{feichtenhofer2019slowfast}  & R50 & 24.2 \\
LFB\cite{wu2019long}  & R50-NL & 25.8 \\
Context-Aware RCNN\cite{wu2020context}  & R50-NL & 28.0 \\
AIA\cite{tang2020asynchronous}  & R50 & 28.9 \\
ACAR\cite{pan2021actor} &  SlowFast-R50 & 28.3 \\
\hline
GCM (ours) & R50 & \textbf{29.1} \\ 
\hline
\hline
\end{tabular}}
\caption{Comparison with the state-of-the-art on \textbf{AVA} v2.1.}
\label{tab:ava21}
\end{table}

\begin{table}[t]
\centering
\footnotesize
\resizebox{\linewidth}{8mm}{

\begin{tabular}{c|c|c|c}
\hline
\hline
\textbf{Approach}  & \textbf{Backbone} & \textbf{Params} & \textbf{mAP}  \\
\hline
Slowfast\cite{feichtenhofer2019slowfast} & R50 & 149MB & 27.34 \\
AIA\cite{tang2020asynchronous} & R50 & 417MB & 29.80 \\
\hline
GCM (ours) & R50 & 264MB & \textbf{30.02} \\ 
\hline
\hline
\end{tabular}}
\caption{Comparison with the state-of-the-art on \textbf{AVA} v2.2. Our approach outperforms the state-of-the-art \cite{tang2020asynchronous}, with much fewer parameters for inference.}
\label{tab:ava22}
\end{table}

\noindent\textbf{Comparisons with state-of-the-art results.} We compare our model with the state-of-the-art methods for action detection on the AVA \cite{gu2018ava} dataset both in version 2.1 and version 2.2. The only difference between the two versions is the accuracy of annotations. Table \ref{tab:ava21} and Table \ref{tab:ava22} report their experimental results. We evaluate these approaches with ResNet-50 backbone for a fair comparison. These state-of-the-arts approaches are all designed for action detection, where ACRN \cite{sun2018actor}, Relation graph \cite{zhang2019structured}, VAT \cite{girdhar2019video}, Context-Aware RCNN \cite{wu2020context}, and AIA \cite{tang2020asynchronous} address the interaction problem between actors and context as well. It can be seen that our model with ResNet-50 achieves the best performance on both datasets, and with much fewer parameters compared to another advanced approach (AIA). The results demonstrate the effectiveness and efficiency of our approach for video action detection. 

\noindent\textbf{Comparison with other interactive blocks.} For further analyzing the effectiveness of our And-Or modules design in action detection, we design a experiment with a specific setting. In this setting, we evaluate the interaction modeling modules with the same inputs and outputs, and only modifying the interactive blocks for fair comparison. The evaluation results are shown in Table \ref{tab:composite}. We observe all the interactive modules gain a better performance than the baseline, and the differences of their performance are quite minimal since all the modules use the same entity features and block setting. Even though, our model still obtains the best performance among these modules both in spatial and temporal space, and requires the least computational resources, which shows the effectiveness and efficiency of our approach.

\begin{table}
\centering
\resizebox{\linewidth}{25mm}{
\begin{tabular}{c|c|c}
\hline
\hline
\multicolumn{3}{c}{w/o long-range temporal modeling} \\
\hline
\hline
\multicolumn{2}{c|}{\textbf{Model}} & \textbf{mAP}  \\
\hline
\multicolumn{2}{c|}{Baseline} & 26.74  \\ 
\multicolumn{2}{c|}{Soft Relation Graph \cite{zhang2019structured}} & 28.18  \\
\multicolumn{2}{c|}{Interaction Block \cite{tang2020asynchronous}} & 28.00  \\
\multicolumn{2}{c|}{Relation Module \cite{santoro2017simple,sun2018actor}} & 28.31  \\
\multicolumn{2}{c|}{GCM w/o long-range modeling} & \textbf{28.78}  \\
\hline
\hline
\multicolumn{3}{c}{w/ long-range temporal modeling} \\
\hline
\hline
\textbf{Model} & \textbf{FLOPs} & \textbf{mAP} \\
\hline
GCM w/o long-range modeling  & 1.00 $\times$ & 28.78 \\
LFB Block \cite{girdhar2019video}  & 4.86 $\times$ & 29.57 \\ 
Interaction Block \cite{tang2020asynchronous}  & 3.76 $\times$ & 29.72 \\
GCM (ours) & 3.69 $\times$ & \textbf{29.76} \\
\hline
\hline
\end{tabular}}
\caption{Evaluation of interaction modeling modules with and without long-range temporal modeling. These modules are implemented by us based on their papers with the specific setting. Our approach demonstrates the effectiveness and efficiency both in short- and long-term interactions.}
\label{tab:composite}
\end{table}

\begin{figure}
    \centering
    \includegraphics[width=1\linewidth]{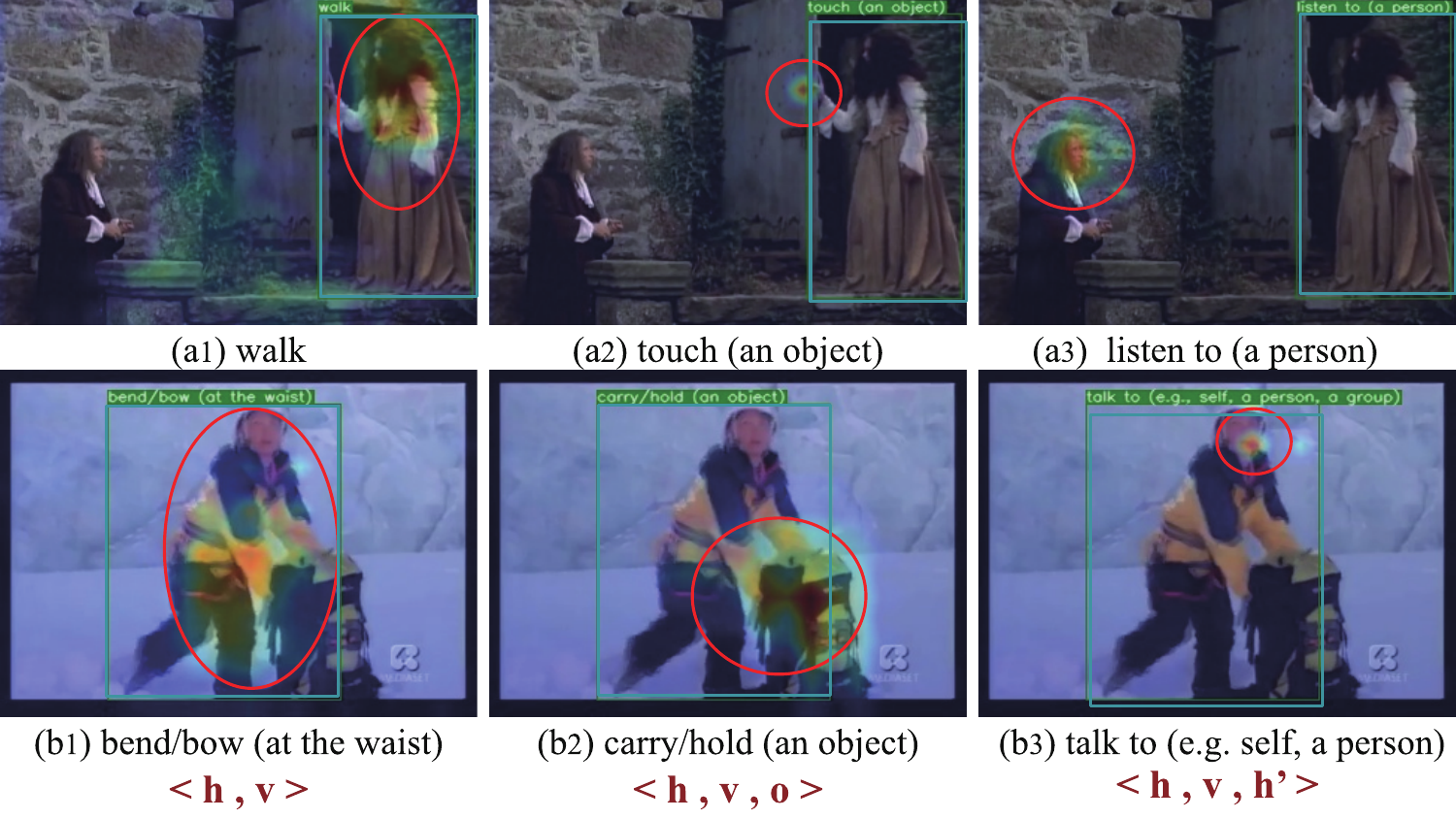}
    \caption{Visualization of heatmaps on AVA, indicating the interesting regions contributing to action predictions,  From left to right: human body movements, human-object interactions, and human-human interactions. The actor is marked in green, and the red circles emphasize the high response scores. The heatmaps vary depending on the categories.}
    \label{fig:visulization}
\end{figure}

\noindent\textbf{Visualization of concurrent actions}. 
To verify the action compositionality of proposed model in a more intuitive way, we show the activation heatmaps on the last convolutional layer using the Grad-CAM \cite{selvaraju2017grad} in Figure \ref{fig:visulization}. It is seen that even for the same actor, the supporting areas varies based on the interactive types. The different concepts of the same actor demonstrate that our model could capture the decomposed elements to some extent.

\begin{table}[t]
\centering
\footnotesize
\begin{tabular}{c|c|c|cc}
\hline
\hline
 & \textbf{Input features} & \textbf{split} & \textbf{top-1} & \textbf{top-5}  \\
\hline
\multirow{4}*{STIN}
~ & Cord & Shuffled & 55.4 & 80.8 \\
~ & Cord & Compositional & 49.7 & 77.3 \\
~ & Cord + OIE & Compositional & 52.1 & 79.8 \\
~ & Cord + OIE + NL & Compositional & 53.1 & 80.6 \\
\hline
\multirow{4}*{GCM}
~ & Cord & Shuffled & 56.8 & 82.0 \\
~ & Cord &  Compositional & 51.3 & 79.0 \\ 
~ & Cord + OIE & Compositional & 52.8 & 80.7 \\ 
~ & Cord + OIE + NL & Compositional & 54.3 & 81.8 \\ 
 \hline
 \hline
\end{tabular}
\caption{Evaluation on \textbf{Something-Else} task. Cord: bounding box coordinates, OIE: Object Identity Embeddings, NL: non-local operators for aggregation. The experimental splits refer to \cite{materzynska2020something}. Our approach outperforms the STIN in all experimental settings.}
\label{tab:something}
\end{table}

\section{Experiments of Something-Else Task}

To validate the generalization ability of our model and compare it with other compositional models, we evaluate our approach in the Something-Else task \cite{materzynska2020something}. The task is built on the Something-Something V2 dataset \cite{goyal2017something} which creates new annotations and splits within it to address compositional action recognition. It contains 174 categories of common human-object interactions, collected via Amazon Mechanical Turk in a crowd-sourced manner. 
We report the results of STIN \cite{materzynska2020something} and our approach in Table \ref{tab:something}. We can see that our proposed GCM outperforms the STIN model in both shuffled and compositional split settings with all types of input features. The result indicates that our approach is effective for modeling the human-object interaction, and has superior generalization capability to compositional action recognition tasks.

\section{Conclusion}

In this paper, we propose a simple, intuitive and effective model for video action detection. We study the composite property of interactive actions and investigate it in a grammatical manner. We construct a novel hierarchical Grammatical Compositional Model for action detection at four semantic levels: a) entities, b) primitive actions, c) concurrent actions, and d) long-range contextual information. The construction of each layer is computed by alternative And and Or node operations in a bottom-up manner. The model can be readily embodied into network modules to empower DNNs with compositionality and provide interpretability by inference parsing.
Experimental results have validated the effectiveness and efficiency of our model.

\bibliography{aaai22}

\end{document}